# Image computing for fibre-bundle endomicroscopy: A review

Antonios Perperidis[1,2], Kevin Dhaliwal[2], Stephen McLaughlin[1], Tom Vercauteren[3]

[1] Institute of Sensors, Signals and Systems (ISSS), Heriot Watt University, EH14 4AS, UK

[2] EPSRC IRC "Hub" in Optical Molecular Sensing & Imaging, MRC Centre for Inflammation Research, Queen's Medical Research Institute (QMRI), University of Edinburgh, EH16 4TJ, UK

[3] School of Biomedical Engineering and Imaging Sciences, King's College London, WC2R 2LS, UK

A.Perperidis@hw.ac.uk (Corresponding Author)

**Abstract**

Endomicroscopy is an emerging imaging modality, that facilitates the acquisition of *in vivo*, *in situ* optical biopsies, assisting diagnostic and potentially therapeutic interventions. While there is a diverse and constantly expanding range of commercial and experimental optical biopsy platforms available, fibre-bundle endomicroscopy is currently the most widely used platform and is approved for clinical use in a range of clinical indications. Miniaturised, flexible fibre-bundles, guided through the working channel of endoscopes, needles and catheters, enable high-resolution imaging across a variety of organ systems. Yet, the nature of image acquisition though a fibre-bundle gives rise to several inherent characteristics and limitations necessitating novel and effective image pre- and post-processing algorithms, ranging from image formation, enhancement and mosaicing to pathology detection and quantification. This paper introduces the underlying technology and most prevalent clinical applications of fibre-bundle endomicroscopy, and provides a comprehensive, up-to-date, review of relevant image reconstruction, analysis and understanding/inference methodologies. Furthermore, current limitations as well as future challenges and opportunities in fibre-bundle endomicroscopy computing are identified and discussed.

## 1. Introduction

The emergence of miniaturised optical-fibre based endoscopes has enabled real-time imaging, at cellular resolution, of tissues that were previously inaccessible through conventional endoscopy. Fibre-bundle endomicroscopy (FBEμ), the most prevalent endomicroscopy platform, has been clinically deployed for the acquisition of *in vivo*, *in situ* optical biopsies in a wide and ever-increasing range of organ systems predominantly in the gastro-intestinal, urological and the respiratory tracts. Customarily, a coherent fibre bundle is guided through the working channel of an endoscope (or a needle) to a region of interest and intravenous or topical dyes are employed to augment tissue fluorescence, enhancing the emitted signal of the imaged structure.

Endomicroscopy has the potential to assist diagnostic and interventional procedures by aiding targeted sampling and increasing diagnostic yield and ultimately reducing the need for histopathological tissue biopsies and any associated delays. To date, the most widespread use of FBEµ (along with fluorescent dyes, such as fluorescein) is in the gastro-intestinal (GI) tract (Fugazza et al., 2016; Wallace and Fockens, 2009; Wang et al., 2015). In particular, FBEµ has been employed in the upper GI tract (East et al., 2016) to detect (i) structural changes in the oesophagus mucosa associated with squamous cell carcinoma and Barrett's oesophagus, and (ii) polyps and neoplastic lesions as well as gastritis and metaplastic lesions in the stomach and duodenum. In the lower GI tract, fibre-endomicroscopy has been utilised to (i) detect colonic neoplasia (Su et al., 2013) and malignancy in colorectal polyps (Abu Dayyeh et al., 2015), as well as to (ii) assess the activity and relapse potential of Inflammatory Bowel Disease (IBD) (Rasmussen et al., 2015; Salvatori et al., 2012).

In pulmonology, the auto-fluorescence (at 488nm) generated through the abundance of elastin and collagen has enabled the exploration of the distal pulmonary tract as well as the assessment of the respiratory bronchioles and alveolar gas exchanging units of the distal lung without the need for exogenous contrast agents. Clinical studies have demonstrated the ability of FBEµ to image a range of pathologies including (i) changes in cellularity in the alveolar space as indicator of acute lunge cellular rejection following lung transplantation (Yserbyt et al., 2014), (ii) cross-sectional and level of fluorescence changes in the alveolar structure in emphysema (Newton et al., 2012; Yserbyt et al., 2017), and (iii) elastic fibre distortion (Yserbyt et al., 2013) and neoplastic changes in epithelial cells (Fuchs et al., 2013; Thiberville et al., 2007; Thiberville et al., 2009) in bronchial mucosa.

Other clinical applications of FBEµ include (but are not limited to) imaging (i) structural epithelial changes observable in bladder neoplasia (Sonn et al., 2009) as well as upper tract urothelial carcinoma (Chen and Liao, 2014), (ii) pancreatobiliary strictures as well as in pancreatic cystic lesions (catheter and/or needle based endomicroscopy), detecting potential malignancy (Karia and Kahaleh, 2016; Smith et al., 2012), (iii) the oropharyngeal cavity, differentiating between healthy epithelium, squamous epithelium and squamous cell carcinoma (Abbaci et al., 2014), and (iv) brain tumours (surgical access), such as glioblastoma, providing immediate histological assessment of the brain-to-neoplasm interface and hence improving tumour resection (Mooney et al., 2014; Pavlov et al., 2016; Zehri et al., 2014). Furthermore, there has been an effort to develop molecularly targeted fluorescent probes, such as peptides (Burggraaf et al., 2015; Hsiung et al., 2008; Staderini et al., 2017), antibodies (Pan et al., 2014) and nanoparticles (Bharali et al., 2005), that can bind and amplify fluorescence in the presence of specific type of tumours (Hsiung et al., 2008; Khondee and Wang, 2013; Pan et al., 2014), inflammation (Avlonitis et al., 2013), bacteria (Akram et al., 2015b) and fibrogenesis (Aslam et al., 2015). Such fluorescent probes will give rise to molecular FBEµ, enhancing the imaging and the diagnostic capabilities of the technology and significantly augmenting utility.

The proliferation of probe-based confocal laser endomicroscopy (pCLE) in clinical practice, along with the emergence of novel, FBEµ architectures and molecularly targeted fluorescent probes necessitate the development of highly sensitive imaging platforms, as well as a range of custom, purpose specific image analysis and understanding methodologies that will assist the diagnostic process. This review provides a brief summary of the available endomicroscopic imaging platforms (Section 2), along with an overview of the state-of-the-art in fibre-bundle endomicroscopic (FBEµ) image computing methods, namely image reconstruction (Section 3), analysis (Section 4) and understanding/inference (Section 5). Owing to its more widespread dissemination, this review paper concentrates in FBEμ image computing. Yet, a small number of relevant image analysis and understanding techniques developed and assessed for other endomicroscopy platforms, offering viable solutions for FBEμ have also been included. Current limitations in FBEμ image computing as well as future challenges and opportunities are also identified and discussed (Section 6).

2. **Technology overview**

To date, four endomicroscopic imaging platforms, all exploiting different fundamental optical imaging technologies, have been commercialised for clinical use (NinePoint, Olympus, Pentax, Mauna Kea, Zeiss). While this review paper concentrates on fibre bundle based systems, a brief description of the currently available endomicroscopic imaging platforms, both commercial and research based, is provided.

NinePoint Medical (Bedford, Massachusetts, USA) developed the NVisionVLE platform, a Volumetric Laser Endomicroscopy (VLE) (Bouma et al., 2009; Vakoc et al., 2007; Yun et al., 2006) device that can acquire in-vivo, high-resolution (7μm), volumetric data of a cavity (e.g. the gastro-intestinal tract) through a flexible, narrow diameter catheter (<2.8mm). VLE combines principles of endoscopic Optical Coherence Tomography (OCT) (Tearney et al., 1997), along with Optical Frequency-Domain Imaging (OFDI) (Yun et al., 2003) to acquire a sequence of cross-sectional images 100-fold faster than conventional OCT, while maintaining the same resolution and contrast. Similar non-commercial OFDI technology has been employed by Tethered Capsule Endoscopy (TCE) to provide an imaging alternative for the gastro-intestinal tract (Gora et al., 2013).

Olympus Medical Systems Co. (Tokyo, Japan) developed a range of prototype endocytoscopes (Hasegawa, 2007), white-light, flexible, contact endoscopes that can image at cellular resolution (up to >1000x magnification). These, now discontinued, prototypes incorporated a miniaturised Charged-Coupled Device (CCD) sensor, the associated objective lenses and an adjacent light source at the distal end of the endoscope. All imaging at the endoscope/tissue contact layer was achieved via light scattering. Olympus provided a variety of options, from (i) full endoscope integration, to (ii) stand-alone, probes that could fit through a 3.7mm endoscope working channel (Ohigashi et al., 2006; Singh et al., 2010). An alternative non-commercial implementation, replacing the miniature sensor with a flexible, coherent fibre bundle enabling imaging at the proximal end of the fibre (Hughes et al., 2013), as well as a range of adaptations to achieve true reflectance endomicroscopy while avoiding back-reflections (Hughes et al., 2014; Liu et al., 2011; Sun et al., 2010) have been proposed.

Confocal laser endomicroscopy (CLE) employs a miniaturised optical fibre to acquire 2D images, predominantly fluorescent, across the examined tissue structure. Inspired from benchtop confocal microscopy (Minsky, 1988), a low-power laser signal (typically at 488nm), focused to a single, finite point within the specimen, is scanned across a two-dimensional imaging plane, generating a 2D image commonly referred to as an optical section. Optical fibres are typically used for relaying light and may act as bi-directional pinholes, rejecting light outside the focal point, reducing the associated image blurring. There are currently, numerous experimental and two commercial CLE platforms with clinical utility, namely endoscope-based (eCLE) and probe-based (pCLE) endomicroscopy. A number of review papers (Jabbour et al., 2012; Oh et al., 2013) provide insight into the current CLE instrumentations. In brief, eCLE integrates a miniaturised confocal scanner into the distal tip of a device using a single core fibre (Delaney et al., 1994; Harris, 1992, 2003). Piezoelectric or electromagnetic actuators can be used to generate the 2D scanning pattern. The tissue signal generated at each individual (scanning) location is transferred through the single-core fibre to a detector and associated processing unit at the proximal end, where the image is accumulated and reconstructed after each complete scan. A clinical eCLE platform was developed by Pentax Medical (Tokyo Japan), integrating the confocal scanning facility into a conventional endoscope. This now discontinued device had a 12.8mm diameter and enabled the acquisition of optical sections at a 500x500µm field of view and 0.7µm lateral resolution. Optiscan (Melbourne, Australia) has recently developed a pre-clinical eCLE platform, comprising of a 4mm diameter standalone micro-endoscope with a <0.5µm lateral resolution (highest commercially available resolution) and a 475x475µm field of view. The acquisition frame-rate in both devices is dependent on the associated acquisition aspect ratio and Z-stack depth (single vs multiple frames), with typically reported values ranging between 0.8 to 6 frames per second (for a single frame). Carl Zeiss Meditec (Jena, Germany) has recently developed a digital biopsy tool for neurosurgery (Leierseder, 2018) based on the underlying Optiscan eCLE technology (475x267µm FOV at 488nm excitation). A number of alternative, non-commercial experimental architectures utilising distal based scanning have been proposed, including (i) high speed imaging (Shi and Wang, 2010) via parallel distal scanning through a multi-core fibre; (ii) dual-axis imaging (Wang et al., 2003), separating the illumination and collection paths and enabling 2D (Liu et al., 2007), 3D (Ra et al., 2008) and multi-colour (Leigh and Liu, 2012) imaging capabilities with improved optical sectioning (axial resolution); and (iii) two-photon imaging (So et al., 2000; Wu and Li, 2010), employing multiple, less energetic photons to induce transition of the imaged fluorescent structure to the desired excitation state, improving the resulting imaging resolution, penetration depth and reducing potential tissue photodamage.

Probe-based confocal laser endomicroscopy (pCLE) utilises a multicore imaging fibre-bundle for the acquisition of 2D optical *en face* sections of a tissue structure. Confocal scanning takes place at the proximal end of the fibre and is relayed by the fibre bundle. Each individual core within the bundle, often combined with a pinhole, rejects light outside the focal plane. Compared to eCLE, the optical setup for pCLE results in a substantially smaller distal endomicroscope as well as higher acquisition frame rates. In contrast, the imaging depth is fixed (and smaller) by the distal optics and the lateral resolution is determined (limited) by the inter-core distance of a particular multicore fibre bundle and distal optics design. (Gmitro and Aziz, 1993; Rouse et al., 2004; Sabharwal et al., 1999) proposed an early implementation of pCLE with (Dubaj et al., 2002; Le Goualher et al., 2004b) introducing refinements for real-time confocal scanning in biomedical applications. Mauna Kea Technologies (Paris, France) developed the Cellvizio pCLE imaging platform along with a wide range of compatible multi-core fibre probes with diameters as small as 0.3mm, and, excluding magnification from distal optics, approximate lateral resolution of 3.3µm and field of views between 300 and 600µm. Additional distal optics can be used to improve the imaging resolution to approximately 1µm, at the expense of field of view (240µm) and diameter (<3mm). The probes' miniature sizes enable use through the working channel of most commercially available endoscopes as well as some needles/catheters, while the relevant high data acquisition rate (>12fps) enables real-time imaging of moving structures, resulting in Cellvizio (pCLE) being the most widely used endomicroscopy platform approved for clinical use. A rapidly growing volume of alternative, non-commercial experimental architectures is being developed, including (i) right-angle stage attachment for standard desktop confocal microscopes (ii) line-scanning confocal endomicroscopy (Hughes and Yang, 2015, 2016), improving the acquisition frame rate without compromising substantially image quality; (iii) flexible and low-cost endomicroscopy architectures (Hong et al., 2016; Krstajić et al., 2016; Pierce et al., 2011; Shin et al., 2010), employing LED, widefield illumination; (iv) structured illumination endomicroscopy (Bozinovic et al., 2008; Ford et al., 2012b; Ford and Mertz, 2013), providing out-of-focus background rejection without beam scanning;

(v) oblique back-illumination endomicroscopy (Ba et al., 2016; Ford et al., 2012a; Ford and Mertz, 2013), collecting phase-gradient images of thick scattering samples; and (vi) multi-spectral imaging (Bedard and Tkaczyk, 2012; Cha and Kang, 2013; Jean et al., 2007; Krstajić et al., 2016; Makhlouf et al., 2008; Rouse and Gmitro, 2000; Vercauteren et al., 2013; Waterhouse et al., 2016). These alternative architectures, along with pCLE can be grouped under the term Fibre-Bundle Endomicroscopy (FBEµ), which, due to its more widespread dissemination, is the technology primarily deliberated throughout this study.

## 3. Image reconstruction

The nature of image acquisition through coherent fibre bundles is a source of inherent limitations in FBEµ imaging. Coherent fibre bundles are comprised of multiple (>10.000) cores that (i) have variable size and shape, (ii) are irregularly distributed across the field of view, (iii) have variable light transition properties, including coupling efficiency and inter-core coupling spread, and (iv) have spatiotemporally variable auto-fluorescent (background) response at certain imaging wavelengths. Such properties directly limit the imaging capabilities of the technology. There has therefore been a substantial interest in the development of effective and efficient approaches to reconstruct FBEµ images, attempting to compensate for these inherent limitations. Table 1 provides an overview of the most relevant image reconstruction studies applicable to FBEµ, while Fig. 1 provides characteristic examples of the associated imaging limitations.

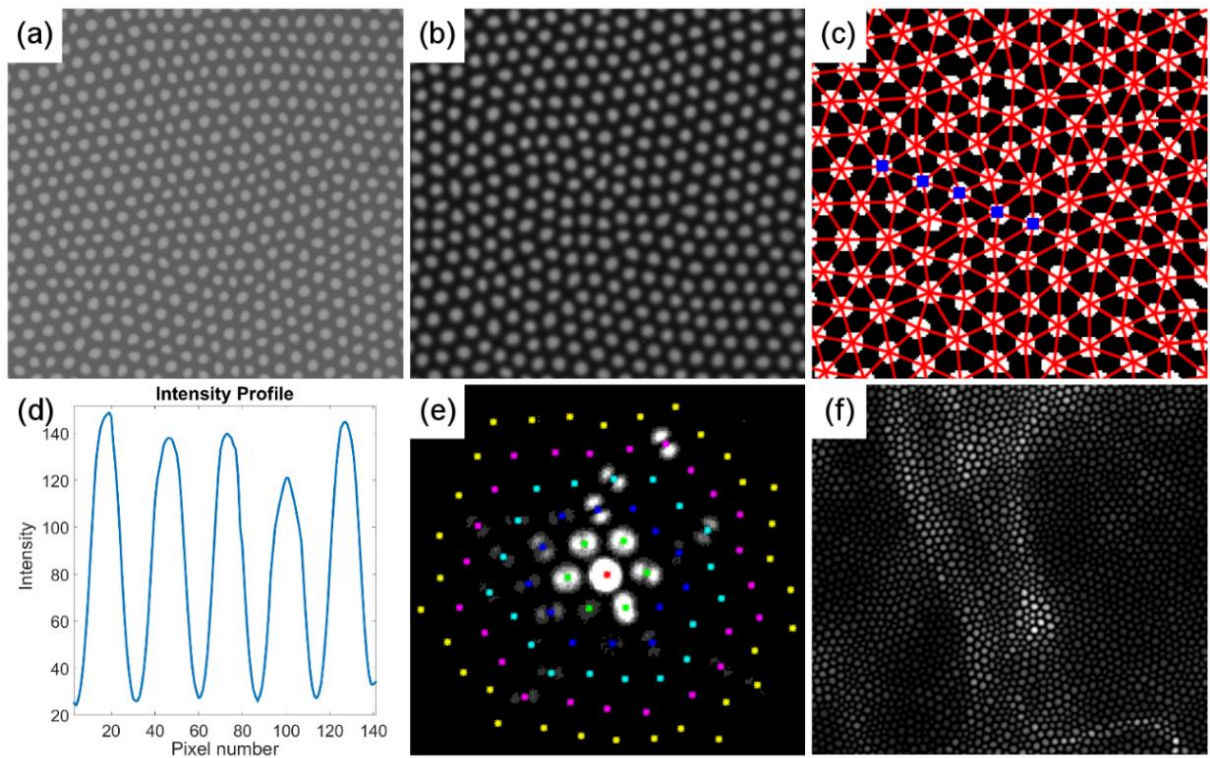

**Fig. 1.** Examples illustrating properties of coherent fibre bundles that limit the imaging capabilities of fibred endoscopy. (a) Scanning Electron Microscopy (SEM) image of commercial coherent fibre bundle (FIGH-30-650S, Fujikura), along with (b) a uniform, flood illumination (520nm) image of the same fibre bundle, using widefield endomicroscopy (Krstajić et al., 2016). The variable size and shape, as well as irregular distribution of the cores is apparent in both (a) and (b). (c) Binary mask and associated Delaunay triangulation of cores, identified within a uniformly illuminated image, similar to (b). (d) Intensity profile across the five cores highlighted in (c) illustrating the variations of coupling efficiency amongst different neighbouring cores. (e) Example inter-core coupling spread at 520nm, as measured by (Perperidis et al., 2017b). (f) Example raw widefield endomicroscopy image of auto-fluorescent alveoli structures from an ex-vivo, human lung. The imaged structured is heavily corrupted by the intrinsic characteristics of the imaging fibre bundle, highlighting the need for effective image reconstruction approaches. Images (a-b), (c) and (e) have been reproduced (cropped) from Figures 6, 7 and 8 respectively of "Characterization and modelling of inter-core coupling in coherent fiber bundles" by (Perperidis et al., 2017b) under CC BY 4.0.

## 3.1. Honeycomb effect

The most visually striking, and limiting artefact, arising from the transmission of the imaged scene through a coherent fibre bundle, is the so-called honeycomb effect. The honeycomb effect, illustrated in Fig. 1, is a consequence of the light being guided from the distal to the proximal end of the individual cores comprising the fibre-bundle but not through the surrounding cladding. Each core, while usually imaged across multiple pixels, contains intensity information on a single, discrete position within the imaged scene. Consequently, the resulting raw image data is a high-resolution rectangular matrix representation of a low-resolution, irregularly-sampled scene. Several studies have attempted to supress/remove the honeycomb effect in fibred endoscopy, generating continuous, high-resolution image sequences.

Throughout the years, a number of approaches employing band-pass filtering in the Fourier domain have been proposed (Dickens et al., 1998; Dickens et al., 1997, 1999; Dumripatanachod and Piyawattanametha, 2015; Ford et al., 2012b; Han et al., 2010; Lee and Han, 2013b; Maneas et al., 2015; Rupp et al., 2007; Winter et al., 2006). Band-pass filters employing a range of different kernels, static and adaptive (derived from the core distribution across the bundle) were typically combined with a range of pre- and post-processing approaches to enhance the performance of suppressing the core honeycomb pattern. Band-pass filtering provides a simple and efficient approach to suppress/remove the honeycomb structure from fibred endoscopic images. However, given the irregularly distributed cores in most modern miniaturised fibrescopes, identifying suitable thresholds in the frequency domain that would remove the honeycomb effect (usually high frequency component) without blurring the underlying imaged structure (usually lower frequency component) can be inherently challenging.

In contrast to band-pass filtering, interpolating amongst the irregular core lattice effectively removes the undesired honeycomb structure while retaining the original image content at the core locations. To accurately identify the locations of each individual core, a uniformly illuminated calibration image is required. Local maxima and the Circular Hough Transforms (CHT) are amongst the wide range of off-the-shelf solutions for identifying core locations. Suggested interpolation methods (Elter et al., 2006; Le Goualher et al., 2004a; Rupp et al., 2007; Rupp et al., 2009; Vercauteren et al., 2006) include (i) $C^0$ continuous nearest neighbour, triangulation-based and natural-neighbour based linear interpolations, (ii) $C^1$ continuous Clough-Tocher interpolation (Amidror, 2002) and a Bernstein-Bezier extension to natural neighbours (Farin, 1990), and (iii) $C^2$ continuous Radial basis functions (Amidror, 2002), b-spline approximation (Lee et al., 1997) and recursive Gaussian filter (Deriche, 1993) adaptation of Shepard's interpolation. (Zheng et al., 2017) attempted to refine the results of a bilinear interpolation using a rotationally invariant adaptation of Non-Local Means (NLM) filters. Moreover, (Winter et al., 2007) proposed an extension of the core interpolation approach for single chip colour cameras, suppressing any false colour moiré patterns. While higher order continuity generated smoother images, a property that can be desirable in particular applications, the associated reconstruction accuracy was shown (Rupp et al., 2009) to be only marginally superior to simple $C^0$ algorithms. On the other hand, for simple Voronoi Tessellation based approaches, all calculations could be performed once, at the calibration stage, generating look up tables to be employed during the subsequent image reconstruction task. Consequently, by generating comparable results with less computational complexity, makes such linear interpolation approaches more attractive candidates for real-time applications.

**Table 1.** Overview of reconstruction approaches for fibred endoscopic imaging.

| Topic | References | Methodology | Comments |
|---|---|---|---|
| Honeycomb effect & Fourier domain filters | (Dickens et al., 1998; Dickens et al., 1997, 1999) | Manual band-reject filters with "high-boost" filter. | Simple to implement, and computationally efficient approaches that suppress the honeycomb structure. |
| | (Han et al., 2010) | Histogram equalisation with Gaussian low-pass filter. | |
| | (Rupp et al., 2007; Winter et al., 2006) | Low pass filter using alternative (circular, star-shaped), rotationally invariant kernels. | However, inherently susceptible to blurring the imaged structures. |
| | (Lee and Han, 2013b) | Gaussian based, notch reject filter, eliminating periodic, high-frequency components. | |
| | (Ford et al., 2012b) | Iteratively blurring (low pass) cladding regions while maintaining core intensities. | |
| | (Dumripatanachod and Piyawattanametha, 2015) | Efficient with 2 1D top-hat filters (equivalent to square kernel). | |
| Honeycomb effect & core interpolation | (Elter et al., 2006; Le Goualher et al., 2004a; Rupp et al., 2007; Rupp et al., 2009) | $C^{0-2}$ continuous interpolation methods on irregular core lattice. | Simple and efficient approaches capable on maintaining the original core information. |
| | (Zheng et al., 2017) | Enhancement of interpolated (bilinear) images using rotationally invariant Non-Local Means. | Successfully employed in clinical/commercial systems. |
| | (Winter et al., 2007) | Correcting for variable core PSF overlap over a colour-sensor's Bayer pattern, suppressing false colour moiré patterns. | |
| Honeycomb effect & image superimposition | (Rupp et al., 2007) | Integrate the core locations of 4 shifted and aligned images, interpolate revised grid. | Capable of removing the honeycomb structure and increase the effective resolution of the acquired data. |
| | (Kyrish et al., 2010; Lee and Han, 2013a; Lee et al., 2013) | Compounding images shifted (translation stage) with a range of predetermined patterns. | |
| | (Cheon et al., 2014a, b; Vercauteren et al., 2006; Vercauteren et al., 2005) | Aligning (compensating for random movements) and combining consecutive frames. | Developing real-time elastic registration approaches is a major challenge. |
| Honeycomb effect & iterative reconstruction | (Han and Yoon, 2011) | Maximising the posterior probability in a Bayesian framework (Markov Random Fields). | Preliminary studies, successful at removing honeycomb structures. |
| | (Liu et al., 2016) | $l_1$ norm minimisation (using iterative shrinkage thresholding - IST) in the wavelet domain. | Not necessarily improve reconstruction error to interpolation. |
| | (Han et al., 2015) | Efficient, non-parametric iterative compressive sensing for inpainting cladding regions. | Computationally costly due to iterative nature. |
| Variable coupling & background response | (Ford et al., 2012b; Le Goualher et al., 2004a; Zhong et al., 2009) | Affine intensity transform incorporating dark and bright-field information at each core. | Capable of supressing (in real time) the effect of spatio-temporally variable coupling and background response. |
| | (Savoire et al., 2012) | Blind calibration exploring neighbouring core correlation to recursively (online) derive gain and offset coefficients in each core. | |
| | (Vercauteren et al., 2013) | Multi-colour extension of (Le Goualher et al., 2004a) dealing with geometric and chromatic distortions. | Successfully employed in clinical/commercial systems. |
| Cross coupling | (Perperidis et al., 2017a) | Quantifying (and integrating into a linear model) cross coupling within fibre bundles. | Effective in supressing the effect of cross coupling. |
| | (Karam Eldaly et al., 2017) | Deconvolution and image reconstruction, reducing the effect of inter-core coupling. | Computationally costly for real-time scenarios. |

Superposition or compounding of spatially misaligned and partially decorrelated images is an approach that has been effective in the enhancement of medical ultrasound images (Perperidis et al., 2015; Rajpoot et al., 2009). In fibred endoscopy, movement of the fibre tip in successive frames accommodates the acquisition of information from the regions previously masked by the fibre cladding. Hence, by effectively aligning the imaged structures and combining a sequence of shifted frames (i) the honeycomb structure can be suppressed/eliminated, (ii) the imaging resolution can be increased. Numerous attempts have examined the effect of different shift patterns, altering the location of the core pattern with respect to the imaged structure, and superposition methods, such as deriving the average or maximum intensity of the aligned images on fibreoscopic images (Kyrish et al., 2010; Lee and Han, 2013a; Lee et al., 2013; Rupp et al., 2007). Alternatively, (Cheon et al., 2014a, b; Vercauteren et al., 2006; Vercauteren et al., 2005) employed the random movements during data acquisition, as would be expected in a realistic clinical scenario, to create an enhanced composite image. The approach was first introduced as part an image mosaicing framework (see Section 4.1) with the main effort being placed in devising an accurate and efficient approach for the alignment of consecutive images. While small translational movements can be efficiently and potentially accurately estimated in real-time, in a realistic scenario with, image distortions as well as elastic and sometimes large structural deformations between consecutive frames, the accurate real-time alignment and compounding can be an eminently challenging task. Increased acquisition frame rate can potentially reduce the deformations between consecutive frames, making their effective alignment more realisable. However, increasing the acquisition frame rate can have detrimental effect in the signal to noise ratio and associated imaging limits of detection.

Recently, several more "sophisticated", iterative methods have been proposed for the reconstruction of fibred endoscopic images and the removal of the associated fixed honeycomb pattern. (Han and Yoon, 2011) employed a Bayesian approximation algorithm to decouple the honeycomb effect. (Liu et al., 2016), based on the empirical observation that natural images tend to be sparse in the wavelet domain, employed $l_1$ norm minimisation in the wavelet domain to remove the honeycomb pattern. (Han et al., 2015) employed an efficient, non-parametric iterative compressive sensing technique for inpainting the cladding regions, without the need of any prior information with regards to the underlying core structure. Limited evaluation (on USAF resolution targets and some biological data) demonstrated the potential of such iterative approaches in image reconstruction, removing the honeycomb artefact as well as fibre bundle defects, while maintaining the spatial resolution and considerably increasing the image contrast and contrast to noise ratio (CNR). However, the current algorithm implementations are considered computationally expensive, making them unsuitable for real time applications. Nevertheless, accelerated, parallel processing though state-of-the-art Graphical Processing Units (GPU) could potentially enable the real-time implementation of such iterative approaches.

## 3.2. Variable coupling and background response

Coherent fibre bundles comprise of a large number cores, commonly in excess of 5000. To reduce the effect of inter-core coupling, neighbouring cores tend to be of variable size and shape. A consequence of this core irregularity is the variable coupling efficiency observed across the fibre bundle. Furthermore, some imaging fibre bundles have exhibited an intrinsic, background auto-fluorescent response at certain imaging wavelengths (e.g. 488nm). Auto-fluorescence, as with coupling efficiency, is also associated with the shape and size of each individual core. These innate fibre properties have a detrimental effect in imaging quality. Consequently, explicit calibration procedures have been developed in an attempt to supress their effect in fibred endoscopic imaging. (Le Goualher et al., 2004a) proposed an off-line calibration process, utilising (i) an image of the fibre auto-fluorescent background (dark-field), as well as (ii) an image of a uniformly fluorescent medium (bright-field). More specifically, for every frame during data acquisition, geometric distortions caused by the resonant scanning mirrors were compensated and the intensity at each core location was normalised using an affine intensity transformation combining the dark and bright-field information. (Ford et al., 2012b; Zhong et al., 2009) extended the (Le Goualher et al., 2004a) approach, introducing additional normalisation terms to partially compensate for camera bias, ambient background light and occasional system realignment. (Vercauteren et al., 2013) adapted the off-line calibration approach in (Le Goualher et al., 2004a) to deal with the distortion compensations (geometric and chromatic) for multi-colour acquisition. In particular, chromatic distortions were estimated and compensated by a symmetric and robust version of the Iterative Closest Point algorithm relying on orthogonal linear regression.

The aforementioned studies assumed constant gain (coupling efficiency) and offset (background auto-fluorescence) for each individual core. However, medium-dependent and slow time-varying coefficient deviations can introduce a static noise pattern on the acquired images. (Savoire et al., 2012) explored the high correlation of signals between neighbouring cores to develop a blind on-line calibration process. For every core in the bundle, (i) linear regression on a temporal window estimated the relative gain and offset coefficients for the associated neighbouring core-pairs, (ii) regularised inversion derived the core's actual gain and offset parameters. To compensate for slow time-varying coefficient changes, the process was performed recursively over temporal windows sufficiently large to enable a robust inversion process.

### 3.3. Inter-core coupling

Inter-core coupling is another limitation in coherent fibre bundles, resulting in blurring of the imaged structures and consequently a worsening in the associated limits of detection in FBEµ. Inter-core coupling has been studied both experimentally (Chen et al., 2008; Wood et al., 2017) and within the theoretical framework of coupled mode theory (Ortega-Quijano et al., 2010; Reichenbach and Xu, 2007; Wang and Nadkarni, 2014), providing (i) insights on the factors affecting cross talk, and (ii) solutions/recommendations for optimal design, selection and optimisation of fibre bundles. Yet, due to the trade-off between cross coupling and core density, cross coupling can be suppressed yet not eliminated through optimal fibre design. In a recent study, (Perperidis et al., 2017a), introduced a novel approach for measuring, analysing and quantifying cross coupling within coherent fibre bundles, in a format that can be integrated into a linear model of the form $v = Hu + w$ with $v$ being the recorded image, $u$ the original signal, $H$ the convolution operator modelling the spread of light, and $w$ an additive observation noise. (Karam Eldaly et al., 2017) employed this linear model and demonstrated the potential of both optimisation-based and simulation-based approaches in reconstructing FBEµ data and reducing the effect of inter-core coupling. However, the computational requirements of the proposed methodology limit their current suitability in real-time clinical applications.

### 4. Image analysis

Analysis of the acquired data and quantification of the imaged structures and potential pathologies is an imperative component to the development of Computer Aided Diagnosis (CAD) systems. Such systems can capitalise on the real-time, optical biopsy capabilities of the technology. Yet, the underlying imaging technology, along with the nature of the clinical data acquisition, generating a steady stream of high resolution images with constricted Field of View (FOV), impose a series of inherent restrictions/challenges to the development of image analysis methodologies. To date, image analysis research for FBEµ can be broadly categorised into (i) mosaicing (Table 2), and (ii) quantification (Table 3) methods. However, the literature appears to be heavily unbalanced, concentrating predominantly on the task of mosaicing frame sequences to extend the associated FOV.

### 4.1. Mosaicing

The miniaturisation of imaging fibre bundles in FBEµ constraints the effective field of view (potentially <500µm) and thus limits sampling diversity, which have implications for navigation, target tissue identification and scene interpretation. To address these inherent limitations, multiple, partially overlapping frames acquired over time can be aligned and combined (stitched) into a single frame with extended field of view. The process has been referred to as image mosaicing. Over the years, there has been considerable research in the development of image mosaicing approaches (Ghosh and Kaabouch, 2016), employed in a range of applications, including endoscopic imaging (Bergen and Wittenberg, 2016). Yet, generic mosaicing approaches do not deal with the inherent properties and limitations of endomicroscopy as described in (Vercauteren 2006). Notably, FBEµ is a direct contact imaging technique. The interaction of a moving rigid fibre-bundle tip with soft tissue may result in non-linear deformations of the imaged structures. A model of probe-tissue interaction for FBEµ was proposed in (Erden et al., 2013). Furthermore, in laser scanning based FBEµ platforms, an input frame does not represent a single point in time. Instead, each sampling point is acquired at a slightly different point in time, resulting in potential deformations when imaging fast moving objects. Finally, the imaged tissue structures are sampled through a sparse, irregularly distributed fibre bundle. Hence, due to these non-linear deformations, motion artefacts and irregular sampling of the input frames, there has been a substantial research interest in the development of custom mosaicing approaches optimised for endomicroscopic data. The proposed methodologies, some currently used in clinical practice, range from simple real-time, to more intricate post-procedural solutions, for either free-hand and/or robotically driven mosaicing platforms. Table 2 and provides an overview of FBEµ mosaicing techniques and characteristic examples of the derived mosaics.

Early mosaicing approaches were post-procedural and addressed both rigid and elastic deformations. (Vercauteren et al., 2006; Vercauteren et al., 2005) were the first studies to identify the necessity for custom mosaicing approaches in endomicroscopy. (Vercauteren et al., 2006) provided a hierarchical framework of frame-to-reference transformations (on the original, sparsely sampled data) to iteratively derive a globally consistent rigid alignment, while compensating for motion induced distortions, as well as for non-rigid deformations. Scattered data approximation was employed to reconstruct a continuous, regularly sampled image from the sparsely sampled inputs merged into a common reference. The proposed method, currently used as part of Mauna Kea's post-procedural analysis software, was tested on phantom and *in vivo* data producing smooth mosaics with extended field of view and enhanced resolution (due to image reconstruction on partially overlapping, irregularly sampled images – see Superposition in Section 3.1). (Loewke et al., 2007a) decomposed the problem into similar components, compensating for global (rigid) as well as local (elastic) transformations (incorporating the effect of motion distortions), maximising the certainty of the registration, both global and local, as an integrated optimisation problem. Averaging of overlapping pixels as well as multi-resolution pyramid blending were tested on both simulated as well as in-vivo data, producing mosaics with smooth image transitions and sharp edges across the imaged structures. Finally, (Hu et al., 2010) adopted a different approach, employing elastic registration of consecutive frames based on optical flow of robust image features and blending the mosaiced frames into a super-resolved image through a Maximum a Posterior (MAP) estimation technique. However, very limited FBEµ data (1 mosaic) were provided for the assessment of the technique.

**Table 2.** Overview of mosaicing approaches for fibred endoscopic imaging.

| Topic | References | Methodology | Comments |
|---|---|---|---|
| Image based, real-time | (Bedard et al., 2012; Vercauteren et al., 2008) | Local rigid alignment through normalised cross correlation matching. | Simple, local, rigid registration based on image similarity maximisation. |
| | (Loewke et al., 2008) | Local rigid alignment through feature based optical flow refined via gradient descent on normalised cross correlation. | Provide valuable real-time feedback during data acquisition for effective mosaic generation. Certain assumptions and model simplifications required for achieving real time performance. |
| Image based, post-procedural | (Vercauteren et al., 2006; Vercauteren et al., 2005) | Hierarchical framework of frame-to-reference transformations (on the original, sparsely sampled data) to derive a globally consistent rigid alignment, while compensating for motion distortions, elastic deformations. Global alignment seen as an estimation problem on a Lie group. | More complex models dealing with a range of local and global, rigid and elastic image transformations. Post-procedural approaches with real-time capacity compromised due to the underlying complex registration models. |
| | (Loewke et al., 2011; Loewke et al., 2007a) | Compensating for global (rigid) as well as local (elastic) transformations (including motion distortion. Fixed correspondence between images were replaced with a Gaussian Potential representing the amount of certainty in the registration. Global and local deformation potentials were maximised in an integrated optimisation problem | |
| | (Hu et al., 2010) | Elastic registration of consecutive frames based on optical flow of robust image features (RANSAC strategy on Lucas-Kanade tracker. A Maximum a Posterior (MAP) estimation based image blending generated super-resolved images. | |
| Image based, dynamic imaging | (Mahé et al., 2015) | Dynamic mosaic obtained by solving a 3D Markov Random Field. Two-stage approach, of static mosaicing followed by stitching of the associated video segments. | Generating mosaics that maintain temporal information in the form of infinite loops. |
| External input based | (Loewke et al., 2007b) | Initial rigid alignment using feedback from a robotic arm determining the five degree-of-freedom position/orientation of the fibre tip. | Actuators/sensors provide feedback on the scanning path improving the efficiency and/or the robustness of mosaicing. Hardware additions are limiting their suitability for endoscopic applications. |
| | (Vyas et al., 2015) | Initial rigid alignment using feedback from a six degree-of-freedom electromagnetic sensor positioned at the tip of the fibre-bundle. | |
| | (Mahé et al., 2013) | Weak a-priori knowledge of the trajectory (spiral scan) used to derive spatio-temporal associations within the frame sequence. A hidden Markov model notation and a Viterbi algorithm was recovered the optimal frame associations, feeding a modification of the mosaicing algorithm by (Vercauteren et al., 2006) to estimate the optimal transform. | |

The complex and descriptive models employed by the preceding methodologies have a direct effect on their computational requirements and consequently their real-time capability. Real-time mosaicing can provide much needed feedback, guiding the data acquisition process, ensuring a smooth continuous path over the desired region of interest. (Vercauteren et al., 2008) proposed an early real-time mosaicing algorithm, integrating translation and distortion (due to finite scanning speed) to a single rigid transformation (estimated through a fast, normalised cross correlation matching algorithm) followed from a simple "dead leaves" model for image blending. A very similar approach of aligning consecutive frames was employed by (Bedard et al., 2012). (Loewke et al., 2008) adopted a two-stage pair-wise registration between consecutive frames, (i) obtaining an initial translation estimate through optical flow on easily trackable features, and (ii) refining the estimate through a gradient descent on cross correlation approach. A multi-resolution pyramid blending algorithm was also employed recombining overlapping regions to a composite image. To achieve real-time performance, these approaches needed to make certain assumptions and hence be subjected to a number of inherent limitations, such as the inability to compensate for global, accumulative alignment errors as well as any elastic deformations. A potential solution to these limitations, which has been adopted by both (Loewke et al., 2011) and by Mauna Kea Technologies, is employing a two-stage mosaicing strategy, including real-time mosaicing for live image acquisition, followed from a more accurate, post-procedural reconstruction.

Most mosaicing approaches for FBEµ have assumed a roughly static scene imaged with a moving field of view. However, this is not always the case in clinical applications with mosaicing removing dynamic information that can be of potential clinical use. (He et al., 2010) proposed a method compensating for a range of movements, both operator induced as well as due to respiratory and cardiac motions, stabilising the field of view for improved monitoring of dynamic structural changes. In (Mahé et al., 2015) dynamic video sequences on static mosaics were integrated, enabling the FOV extension without the associated loss of dynamic structural changes throughout the acquisition. A two-stage approach, of static mosaicing followed by stitching of the associated video segments was employed to reduce computational load. Visual artefacts at the seams across the mosaic were suppressed using a gradient-domain decomposition. Dynamic mosaics (infinite loops) from six organs (oesophagus, stomach, pancreas, bladder, biliary duct and colon) with various conditions were produced and clinically assessed by four experts. The produced visual summaries indicated higher level of consistency with the original data compared to static mosaicing.

Mosaicing techniques for FBEµ have, for the most part, concentrated in aligning and blending images with no a priori information on the acquisition trajectory, based exclusively on topology inference through the changes on the imaged structures. Such approaches call for large overlap amongst adjacent frames and, for effective, smooth results, can be computationally expensive. There has therefore been interest in incorporating such a priori trajectory information in the mosaicing process. (Loewke et al., 2007b) utilised feedback from a robotic arm determining the five degree-of-freedom position and orientation of its end-effector (along with projective geometry) as an initial global rigid alignment amongst a frame sequence. (Vyas et al., 2015) replaced the robotic arm with a six degree-of-freedom electromagnetic sensor positioned at the tip of the fibre-bundle in a proof-of-principle study. The positioning feedback from the sensor acted as a coarse global alignment followed from a fine tuning similar to (Vercauteren et al., 2008). In (Mahé et al., 2013) they used weak prior knowledge of the trajectory (spiral scan) to derive spatio-temporal associations within the frame sequence, linking overlapping frames from successive branches of the spiral scan and estimating optimal transforms similar to (Vercauteren et al., 2006). While these approaches have been reported to improve the efficiency and/or the robustness of the mosaicing process, they require additional actuators/sensors at the tip of the fibre bundle, either to drive or to provide feedback on the scanning path. Such hardware additions are currently limiting their suitability for endoscopic applications, necessitating for future miniaturisation of the relevant technologies.

The previous studies described above dealt predominantly with free hand movements for producing an extended field of view through image mosaicing. While free-hand mosaicing is a very valuable tool, the ability to create customised scanning paths, ensuring a full imaging coverage of the region of interest, is also highly desirable. A number robotised distal scanning tips have been proposed (Erden et al., 2014; Rosa et al., 2013; Rosa et al., 2011; Zuo et al., 2015; Zuo et al., 2017b) facilitating customised and structured mosaic acquisition paths (such as spiral and raster scans) for FBEµ. While a diverse set of architectures has been proposed, the plurality of solutions catered mostly for Minimal Invasive/Laparoscopic Surgery applications, with some prototypes suitable for endoscopic applications (Zuo et al., 2017a). Further miniaturisation is therefore imperative for facilitating robotically controlled scanning in endoscopy. In robotised scanning, the complex 3D surface of many of the examined structures, along with the lack of haptic feedback may result in loss of contact, or excessive contact/pressure between the fibre bundle and the imaged structure. (Giataganas et al., 2015) created an adaptive probe mount that could maintain constant (low force magnitude) contact between the tissue and the imaging probe. Furthermore, the direct contact of the hard tip of the fibre bundle with the soft tissue can lead to tissue deformations, resulting in accumulative deviation off the desired path throughout a scan. There have been studies attempting to understand this soft tissue behaviour and provide feedback to the robotised scanner in an attempt to compensate for the anticipated path deviations. The feedback can be (i) through determining the loading-distance prior an automated scan and compensating respectively by adjusting the scan path (Erden et al., 2013), or (ii) through visual servoing (Rosa et al., 2013), estimating the imaged path in real-time through the mosaiced image data and adjusting accordingly to meet the desired scan path. While further research and development, especially in their miniaturisation is necessary, robotic scanning is anticipated to serve as a key milestone to the adoption of image mosaicing in a wide range of clinical endoscopic and laparoscopic procedures. Yet, the detailed discussion on such robotised scanning approaches is beyond the scope of this paper. (Zuo and Yang, 2017) has produced a thorough review on endomicroscopy for robot assisted intervention, providing details and discussion on a wide range of relevant studies.

### 4.2. Quantification

Aside from image mosaicing, there have been a limited number of image analysis studies for FBEµ images. These studies have predominantly concentrated on the detection and quantification of particles and structures that can act as indicators of pathological or physiological processes in the circulatory system, oropharyngeal, gastrointestinal and pulmonary tracts. For the most part, empirical, ad hoc observations, combined with simple, off-the-shelf image analysis approaches, have been employed. This section along with Table 3 provide a brief overview of the most relevant image analysis/quantification studies for fibred endomicroscopic data.

In the circulatory system, (Savoire et al., 2004) proposed a method to estimate the velocity of Red Blood Cells (RBC) within micro-vessels from a single endomicroscopic frame, exploiting the skewing artefact introduced on fast moving RBC due to the relatively slow scanning speed of the vertical axis component (resulting in circular

RBCs appearing ellipsoidal). (Perchant et al., 2007) developed algorithms to track and align a region of interest over consecutive frames for cell traffic analysis and blood velocity estimation. (Huang et al., 2013) examined the variability in stained cardiac tissue structures imaged through FBEµ as a means for intraoperatively identifying nodal tissue in living rat hearts with potential application to neonatal open-heart surgery. In the orapharyngeal tract, (Mualla et al., 2014) identified the borders and locations respectively of epithelial cells in the mucosa layer of vocal chords as the first step to analysing and quantifying structural changes.

In the gastrointestinal tract, (Couceiro et al., 2012) developed a methodology that employed off-the-shelf algorithms for segmenting and quantifying intestinal crypts in endomicroscopic images as a potential indicator for Inflammatory Bowel Disease. Similarly, (Prieto et al., 2016) employed crypt detection as a first step towards automated classification between benign and dysplastic epithelial tissue in colorectal polyps. (Boschetto et al., 2015a; Boschetto et al., 2016b; Boschetto et al., 2015b) attempted to semi-automatically analyse and quantify fluorescent endomicroscopic images of the gastro-intestinal mucosa, as a first step to assist diagnosis and monitoring of Celiac Disease. (Boschetto et al., 2016b; Boschetto et al., 2015b) proposed methodologies for segmenting intestinal villi, while (Boschetto et al., 2015a) proceeded in detecting and segmenting cells within the villi and differentiating between columnar and goblet cells of the epithelium.

Finally, in the pulmonary tract, (Namati et al., 2008) analysed mice distal lung images and automatically quantified the number and size of alveolar sacs. (Perez et al., 2017) applied a sequence of off-the-shelf image processing operations to count fluorescently labelled Mesenchymal Stem Cells injected into rat lungs, as a potential indicator for lung repair in radiation induced lung injury. (Karam Eldaly et al., 2018) employed a fully unsupervised, hierarchical Bayesian approach for detecting bacteria labelled with a (green) fluorescent smart-probe (Akram et al., 2015a) within the, highly auto-fluorescent (also green) distal lung. The algorithm was an extension of (McCool et al., 2016) for denoising along with outlier detection and removal in sparsely, irregularly sampled data. Such fully unsupervised approches offer a flexible and consistent methodology to deal with uncertainty in inference when limited amount of data or information is available. (Seth et al., 2017, 2018) quantified bacterial and cellular load in the human lung adopting and adapting a learning-to-count (Arteta et al., 2014) approach, employing a multi-resolution, spatio-temporal template matching scheme using radial basis functions network.

5. **Image understanding**

Another component of the image computing pipeline is the higher-level understanding and exploitation of the acquired, reconstructed and sometimes processed data, in an attempt to extract clinically and biologically relevant information, and consequently guide the diagnostic process. Due to the nature of FBEµ data acquisition in a clinical setup, a large volume of continuous frame sequences is generated, sometimes surpassing 1000 frames per video. These video sequences include uninformative/corrupted frames, off-target frames outside the examined anatomic structure and/or region of interest, as well as a range of on-target frames from healthy and pathological structures. This large, and sometimes very diverse, data volume acts as a major bottleneck in the analysis and quantification of the data, increasing the required human/computational resources, and potentially diluting the objectiveness of associated clinical procedure. The main body of FBEµ image understanding research to date can be broadly categorised into frame (i) classification (Tables 4-5 and Fig. 2), and (ii) content-based retrieval methods (Table 6).

**Table 3.** Overview of quantification approaches for fibred endoscopic imaging.

| Organ (System) | Quantifying | References | Methodology | Comments |
|---|---|---|---|---|
| Circulatory | Red blood cell velocity. | (Savoire et al., 2004) | Thresholding and line-fitting (M-estimators) translated (through trigonometry) to RBC velocity. | Inventive use of known and quantifiable artefact in raster scanning imaging systems for deriving physiological information. |
| | | | | Preliminary results with uncertain clinical relevance. |
| | | (Perchant et al., 2007) | ROI tracking and alignment through (i) scanning distortion compensation, and (ii) global affine registration, for blood velocity estimation through spatio-temporal correlation. | Feasibility study. |
| | | | | Preliminary results with uncertain clinical relevance. |
| Oropharyngeal | Epithelial cells in vocal chords. | (Mualla et al., 2014) | Watershed segmentation (borders) and local minima detection (location). | Empirical, ad-hoc approach employing off-the-shelf image analysis methods. |
| | | | | Limited data can potentially lead to poor generalisation of the proposed methodology. |
| Gastro-intestinal | Intestinal crypts in Inflammatory Bowel Disease. (eCLE) | (Couceiro et al., 2012) | Detecting (local maxima), segmenting (ellipse fitting on edge detection) and quantifying (number, connectivity). | Empirical, ad-hoc approaches employing off-the-shelf image analysis methods |
| | Intestinal crypts in colorectal polyps. | (Prieto et al., 2016) | Contrast enhancement, thresholding (Otsu's) and morphological filters (erosion, centre of mass, circularity). | Heuristic parameter estimation, hard thresholds and limited data can potentially lead to poor generalisation of the proposed methodologies. |
| | Goblet cells in villi. (eCLE) | (Boschetto et al., 2015a) | Detecting (matched filters), segmenting (Voronoi diagrams) cells and identifying (hard threshold) goblet cells within the villi. | |
| | Intestinal villi. (eCLE) | (Boschetto et al., 2015b) | Detect via morphological filters (top-hat, morphological reconstruction and closing) and quad-tree decomposition. | |
| | | (Boschetto et al., 2016b) | Subdivide to superpixels, extract features and classify through Random Forests to generate a binary segmentation map. | Employing established data driven approaches with reasonable size of data, resulting on better generalisation potential. |
| Pulmonary | Alveoli sacs in mice distal lung. | (Namati et al., 2008) | Segmenting (optimum separation thresholding) and quantifying (8-point connectivity) alveolar sacs. | Limited data and uncertain translatability to human alveoli sacks due to their large size relative to the limited field of view. |
| | Stained mesenchymal stem cells in rat lungs. | (Perez et al., 2017) | Contrast stretch, denoise (opening), threshold and count (connected component analysis). | Empirical, ad-hoc approach employing off-the-shelf image analysis methods. |
| | Stained bacteria in distal lung. | (Karam Eldaly et al., 2018) | Outlier detection using a hierarchical Bayesian model along with a MCMC algorithm based on Gibbs sampler. | More elaborate approaches, adopting model-based and data-driven methodologies. |
| | Stained bacteria and cells in distal lung. | (Seth et al., 2017, 2018) | Bacterial and cellular load using spatio-temporal template matching with a radial basis functions network. | They have potential for good generalisation and translation to clinical applications. |

**Table 4.** Overview of classification approaches for fibred endoscopic imaging employing traditional machine learning.

| Organ (System) | Classifying | References | Methodology | Comments |
|---|---|---|---|---|
| Pulmonary | Distal lung alveolar abnormalities. | (Desir et al., 2012a; Désir et al., 2010; Desir et al., 2012b; Hebert et al., 2012; Heutte et al., 2016; Koujan et al., 2018; Petitjean et al., 2009; Saint-Réquier et al., 2009) | Features: First Order Statistics, GLCMs, LBPs, SIFT, Scattering Transform, FREAK, ORB, Homomorphic filters, Structural Information (Canny and Sobel Edge Detectors), Sparse - Irregular LBPs, LQPs, HOGs, LDPs, Homogeneity, Spatial Frequency, Fractal Texture, Intensity, Wavelet and CNN Features. | Simple and effective methodologies performing in most part binary classification. Results are positive indicating the potential strength of simple approaches in classifying endomicroscopic images. |
| | Informative frames within videos. | (Leonovych et al., 2018; Perperidis et al., 2016) | | Primary limitations include (i) the limited scope of the classification, for example health VS pathological, when endomicroscopic sequences contain a plethora of frame classes, and (ii) the limited number of images used for training, testing and evaluation, making the proposed methodologies susceptible to a range of biases. |
| | Cancerous nodules in airways and distal lung. | (He et al., 2012; Rakotomamonjy et al., 2014; Seth et al., 2016) | Classifiers: K-NN, SVMs, SVM-RFE, Gaussian Mixture Models, LDA, QDA, Random Forests, Generalised Linear Model, Gaussian Processes, Boosted Cascade of Classifiers, Neural Networks. | |
| Gastro-intestinal | Oesophagus epithelial changes. | (Ghatwary et al., 2017; Veronese et al., 2013; Wu et al., 2017) | | |
| | Intestinal adenocarcinoma. (eCLE) | (Ştefănescu et al., 2016) | Multiclass: One-vs-all and one-vs-one ECOCs, binary tree classification, Recursive SVM tree and Naïve Bayes. | |
| | Colorectal polyps. | (André et al., 2012b; Zubiolo et al., 2014) | Other: Pruning trees for non-detection; feature selection (i.e. SDA, FSS and PCA) for dimensionality reduction; visual coding (Bag of Words, Sparse Coding and Fisher Kernel Coding) and classification on mosaics for enhanced classification performance. | |
| | Celiac disease. (eCLE) | (Boschetto et al., 2016a) | | |
| Oropharyngeal | Pathological epithelium. | Jaremenko et al., 2015; Vo et al., 2017) | | |
| Brain | Brain tumours (glioma and meningioma). | (Kamen et al., 2016; Wan et al., 2015) | | |
| Ovaries | Epithelial changes | (Srivastava et al., 2005; Srivastava et al., 2008) | | |

## 5.1. Image classification

Classification of frames on pre-determined, clinically defined cohorts based on their content is currently the most investigated area of FBEµ image computing research. An abundance of studies have applied binary as well as multi-class classification on endomicroscopic images of a range of organ systems in an attempt to identify cancer in ovarian epithelium (Srivastava et al., 2005; Srivastava et al., 2008), abnormalities in distal lung alveolar structures (Desir et al., 2012a; Désir et al., 2010; Desir et al., 2012b; Hebert et al., 2012; Heutte et al., 2016; Koujan et al., 2018; Petitjean et al., 2009; Saint-Réquier et al., 2009), informative frames in brain (Izadyyazdanabadi et al., 2017a; Izadyyazdanabadi et al., 2017b) and pulmonary videos (Leonovych et al., 2018; Perperidis et al., 2016), cancerous nodules in the airways (Gil et al., 2017; He et al., 2012; Rakotomamonjy et al., 2014) and distal lung (Seth et al., 2016), pathological epithelium in the oropharyngeal cavity (Aubreville et al., 2017; Jaremenko et al., 2015; Vo et al., 2017), changes in oesophageal epithelium in cases of Barrett's oesophagus (Ghatwary et al., 2017; Hong et al., 2017; Veronese et al., 2013; Wu et al., 2017), adenocarcinoma (Ştefănescu et al., 2016), colorectal polyps (André et al., 2012b; Zubiolo et al., 2014) and celiac disease (Boschetto et al., 2016a) in intestinal epithelium, neoplastic tissue in breast nodules (Gu et al., 2017), as well as two types of common brain tumours, glioblastoma and meningioma (Kamen et al., 2016; Murthy et al., 2017; Wan et al., 2015). Methodologically, most of the aforementioned studies employed the same basic structure, defining a hand-crafted feature space descriptive of the underlying imaged structure, training a range of classifiers, to distinguish between pre-determined frame categories. For organs/structures that do not exhibit any auto-fluorescence at the imaging wavelengths, fluorescence dyes such as methylene blue and fluorescein, and molecular probes (He et al., 2012) were employed to generate the necessary fluorescent signal.

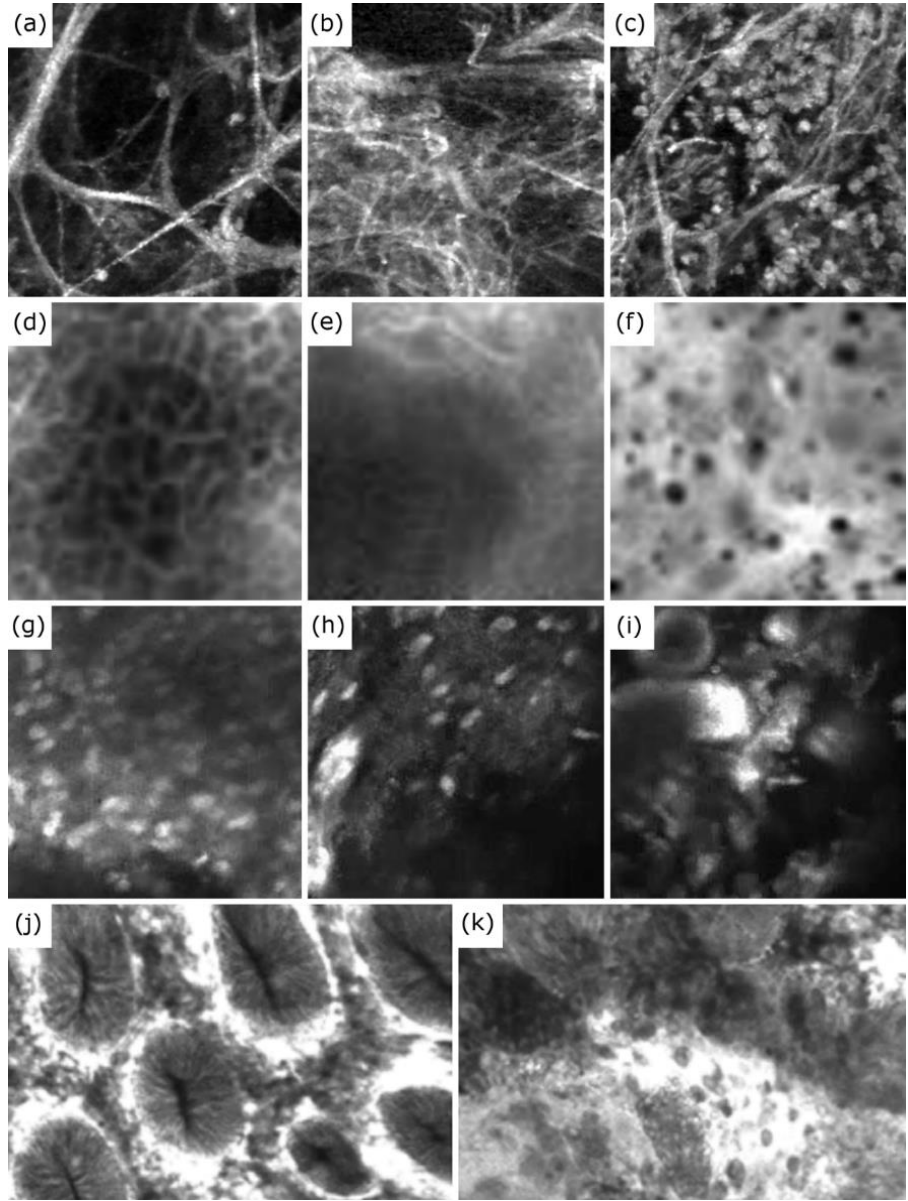

**Fig. 2.** Examples of structural changes observed in OEM images across variety of organ systems and conditions. These structural changes have been used to classify/detect a range of clinically relevant pathologies. (a-c) Difference in tissue structure in the alveoli structures of the distal lung, indicating (a) healthy and (b) pathological elastin strands, as well as (c) alveoli sacs flooded with cells. (d-f) Difference between (a) healthy and (c) cancerous oral epithelium, along with (b) an example of oral epithelium with limited textural information where classification can be challenging. (g-i) Difference between (g-h) Glioblastoma and (i) Meningioma brain tumour images. (j-k) Difference between (j) healthy colon mucosa and (k) adenocarcinoma (Ştefănescu et al., 2016). Images (d-f) have been reproduced (cropped) from Figure 6 of "Automatic Classification of Cancerous Tissue in Laserendomicroscopy Images of the Oral Cavity using Deep Learning" by (Aubreville et al., 2017) under CC BY 4.0. Images (g-i) have been reproduced (cropped) from Figure 3 of "Automatic Tissue Differentiation Based on Confocal Endomicroscopic Images for Intraoperative Guidance in Neurosurgery" by (Kamen et al., 2016) under CC BY 4.0. Images (j) and (k) have been reproduced (cropped) from Figures 2 and 3 respectively of "Computer Aided Diagnosis for Confocal Laser Endomicroscopy in Advanced Colorectal Adenocarcinoma" by (Ştefănescu et al., 2016) under CC BY 4.0.

Commonly used feature descriptors include (i) first order image statistics, (ii) structural information through Skeletonisation, Sobel and Canny Edge Detectors, etc. (iii) Haralick's texture parameters derived through Gray Level Co-occurrence Matrices (GLCM), (iv) Local Binary Patterns (LBP) and their variation of Local Quinary Patterns (LQP), and (v) Scale Invariant Feature Transforms (SHIFT). Other less adopted descriptors employed as discriminative features include (i) spatial frequency based features extracted at Fourier domain (Srivastava et al., 2005; Srivastava et al., 2008), (ii) fractal analysis (Ştefănescu et al., 2016), (iii) Scattering transform (Rakotomamonjy et al., 2014; Seth et al., 2016), (iv) Fast Retina Keypoint (FREAK) (Wan et al., 2015), (v) Oriented FAST and rotated BRIEF (ORB) (Wan et al., 2015), (vi) Histogram of Oriented Gradients (HOG) (Gu et al., 2016; Vo et al., 2017), (vii) textons (Gu et al., 2016), (viii) Local Derivative Patterns (LDP) (Vo et al., 2017), as well as (ix) features extracted from Convolutional Neural Networks (CNN) prior to the fully connected layer employed for computing each class score (Gil et al., 2017; Vo et al., 2017). (Leonovych et al., 2018) introduced Sparse Irregular Local Binary Patterns (SILBP), an adaptation of LBPs taking into consideration the sparse, irregular sampling imposed by the imaging fibre bundle on FBEµ images. Feature spaces combining two or more of the above descriptors are also frequent, with descriptors customarily extracted from the whole image, yet in some cases, regular or randomly distributed sub-windows/patches have been used, either on their own, or in conjunction to the whole image feature space.

**Table 5.** Overview of classification approaches for fibred endoscopic imaging going beyond traditional machine learning.

| Organ (System) | Classifying | References | Methodology | Comments |
|---|---|---|---|---|
| Pulmonary | Cancerous nodules in airways. | (Gil et al., 2017) | Unsupervised classification (compensating for limited data availability) using graph representation and community detection algorithms. | Early FBEµ classification approaches going beyond the traditional machine learning pipeline, exploring methods such as Convolutional Neural Networks (off the self as well as custom), transfer learning, unsupervised learning and multi-modal learning at a latent space.<br><br>The results are very promising. Yet, more data, both in terms of numbers as well as in terms of diversity are necessary. Furthermore, custom solutions, taking into consideration the inherent FBEµ imaging properties, could further enhance the classification performance. |
| Gastro-intestinal | Oesophagus epithelial changes. | (Hong et al., 2017) (Aubreville et al., 2017) | Custom CNN architecture for the multi-class frame classification. | |
| Oropharyngeal | Pathological epithelium. | (Aubreville et al., 2017) | Full-training of LeNet-5 and shallow fine-tuning the Inception v3 (using the ImageNet database). | |
| Brain | Informative frames within videos. | (Izadyyazdanabadi et al., 2017a; Izadyyazdanabadi et al., 2017b) | Fully-trained AlexNet and GoogleNet as well as comparing the between full training and transfer learning through fine-tuning using the ImageNet database. | |
| | Brain tumours. | (Murthy et al., 2017) | Novel Cascaded CNN, discarding easy images at early stages, concentrating on challenging ones at subsequent, expert shallow nets. | |
| Breast | Cancerous breast nodules. | (Gu et al., 2017) | Multi-modal (FBEµ mosaics and histology) classification mapping the original features to a latent space for improved SVM performance. | |

A number of well-established classifiers have been assessed, including (i) k-Nearest Neighbours (kNN) (André et al., 2012b; Desir et al., 2010; Hebert et al., 2012; Saint-Réquier et al., 2009; Srivastava et al., 2005; Srivastava et al., 2008), (ii) Linear and Quadratic Discriminant Analysis (LDA and QDA) (Leonovych et al., 2018; Srivastava et al., 2005; Srivastava et al., 2008), (iii) Support Vector Machines (SVM) and their adaptation with Recursive

Feature Elimination (SVM-RFE) (Desir et al., 2010; Desir et al., 2012b; Jaremenko et al., 2015; Leonovych et al., 2018; Petitjean et al., 2009; Rakotomamonjy et al., 2014; Saint-Réquier et al., 2009; Vo et al., 2017; Wan et al., 2015; Zubiolo et al., 2014), (iv) Random Forests (RF) and variants such as Extremely Randomised Trees (ET) (Desir et al., 2012a; Heutte et al., 2016; Jaremenko et al., 2015; Leonovych et al., 2018; Seth et al., 2016; Vo et al., 2017), (v) Gaussian Mixture Models (GMM) (He et al., 2012; Perperidis et al., 2016), (vi) Boosted Cascade of Classifiers (Hebert et al., 2012), (vii) Neural Networks (NN) (Ştefănescu et al., 2016), (viii) Gaussian Processes Classifiers (GPC), and (ix) Lasso Generalised Linear Models (GLM) (Seth et al., 2016). Most studies employed leave-k-out and k-fold cross validation to assess the predictive capacity of the proposed methodology on limited, pre-annotated frames. In an attempt to enhance the classification performance and/or reduce the computational workload required for training and testing, some studies incorporated additional steps in the classification pipeline. In particular, feature selection (dimensionality reduction in feature space) such as Stepwise Discriminant Analysis (SDA), Forward Sequential Search (FSS), and Principal Component Analysis (PCA) were also used (Perperidis et al., 2016; Srivastava et al., 2005; Srivastava et al., 2008) prior to the classification process. Furthermore, visual coding schemes, such as Bag-of-Words, Fisher Kernel Coding and Sparse Coding (Kamen et al., 2016; Vo et al., 2017; Wan et al., 2015), as well as reduction of non-detection, minimising the incorrectly classified images through rejection mechanisms (Desir et al., 2012a; Heutte et al., 2016), have been investigated.

Classification of endomicroscopic images has predominantly concentrated in binary cases, with a very limited number of studies having attempted multi-class classification (Boschetto et al., 2016a; Ghatwary et al., 2017; Hong et al., 2017; Koujan et al., 2018; Veronese et al., 2013; Wu et al., 2017; Zubiolo et al., 2014). To this end, (Boschetto et al., 2016a) employed a multi-class Naïve Bayes classifier. (Koujan et al., 2018) adopted the One-Versus-All (OVA) Error Correcting Output Codes (ECOC), a popular method (along with other ECOCs such as One-Versus-One and Ordinal) for multi-class classification using binary classifiers. (Ghatwary et al., 2017; Veronese et al., 2013) tackled the multi-class problem as a pre-determined sequence (tree) of binary classifications (through SVM), while (Zubiolo et al., 2014) employed graph theory tools (minimum cut) to recursively estimate the optimal associated bi-partitions (large SVM margin). Hierarchical (tree) binary classifications can potentially reduce the classification complexity from linear for OVA to logarithmic. (Wu et al., 2017) improved the performance of multi-class classification performance incorporating unlabelled images through an adaptation of semi-supervised approach called Label Propagation method introduced by (Zhou et al., 2003).

There have recently been some studies that do not follow the same basic structure of training a classifier on a hand-crafted feature space descriptive of the underlying imaged structures. (Gil et al., 2017) proposed an unsupervised classification approach to compensate for the limited quantity of data available for training and testing decision support systems. The methodology used graph representation to codify feature space connectivity followed by community detection algorithms (Cazabet et al., 2010), representing space topology and detecting associated image communities. (Gu et al., 2016) incorporated features extracted from endomicroscopy mosaics as well as associated histology images, to a supervised framework, mapping the original features to a latent space by maximising their semantic correlation. The derived latent features outperformed mono-modal features in binary classification (SVM) of breast cancer images. Furthermore, recent advances of Deep Learning architectures, such as Convolutional Neural Networks (CNN), have resulted in numerous powerful tools for binary or multi-class image classification, without the need for explicit definition of feature descriptors. (Hong et al., 2017) proposed a custom CNN architecture with for the multi-class classification of epithelial changes in Barrett's oesophagus.

(Aubreville et al., 2017) adopted and adapted two established CNN architectures for the detection of cancerous tissue in the oral cavity, (i) a patch-based classification based on full-training of LeNet-5 (Lecun et al., 1998), as well as (ii) a whole image classification based on shallow fine-tuning the Inception v3 network (Szegedy et al., 2016) pre-trained using ImageNet database (Deng et al., 2009). Similarly, (Izadyyazdanabadi et al., 2017a) fully-trained AlexNet (Krizhevsky et al., 2012) and GoogleNet (Szegedy et al., 2015) for the detection of diagnostic frames in brain endomicroscopy. (Murthy et al., 2017) presented a novel multi-stage CNN, discarding images classified with high confidence at early stages, concentrating on more challenging images at subsequent, expert shallow networks. The proposed network demonstrated substantial improvement on traditional feature/classifier as well as CNN architectures when classifying (binary endomicroscopic brain tumour images. (Izadyyazdanabadi et al., 2017b) compared the classification performance amongst fully training CNNs from scratch against transfer learning through fine-tuning, shallow (fully connected layers) or deep (whole network), of pre-trained networks using conventional image databases such as ImageNet. Similar to (Tajbakhsh et al., 2016), fine-tuning was found to be able to provide better or at least similar classification performance to training from scratch on limited medical image databases.

## 5.2. Image retrieval

While a less prolific research area to the closely related image classification, a number of studies have developed Content Based Image Retrieval (CBIR) frameworks for endomicroscopic data. Unlike image classification, that groups images to a number of pre-determined (trained) classes, CBIR methods search a database to find (and return) the images that are most similar (based to some image extracted feature set) to a given (query) image. In an early attempt, (André et al., 2009a) adapted the Bag of Visual Words (BVW) approach of (Sivic and Zisserman, 2008) to endomicroscopic images, containing discriminative texture information (SIFT) extracted across a regular grid of overlapping disks at various scales (radius). (André et al., 2009b) introduced to the retrieval process (i) spatial relationship between local features by exploiting the co-occurrence matrix of the visual words labelling the local features in each image, as well as (ii) temporal relationship between successive frames in a video sequence, by including image mosaics projecting the temporal dimension onto an extended field of view. In an attempt to avoid computationally costly non-rigid deformations required for a robust mosaic image, (André et al., 2010) proposed a video retrieval approach named Bag of Overlap-Weighted Visual Words (BOWVW). BOWVW computed independently the BVW signatures from individual frames within a video sub-sequence, as per (André et al., 2009a), and weighted the associated contributions (frame overlap rate) of their individual dense local regions to a single signature for the sub-sequence. The subsequence signatures were then incorporated (normalised sum) to a single signature for a whole video. The aforementioned studies were compared and combined to a single, integrated video retrieval approach (André et al., 2011b). However, due to the challenging task of generating ground truth for the evaluation of content based retrieval, the proposed methodology was evaluated as a binary classification task between neoplastic and benign epithelium in Colonic Polyps (André et al., 2012b).

In an attempt to address the challenging evaluation of true retrieval performance (André et al., 2011a) (i) developed a tool for generating the perceived similarity ground truth, enabling the direct evaluation of endoscopic video retrieval, and (ii) employed this ground truth information by employing a similarity distance learning technique to derive an optimal mapping of video signatures, improving the discrimination of similar video pairs. Another challenge in retrieval systems is bridging the "semantic gap" between the (sometimes conflicting) low-level visual features, extracted computationally from the images, and high-level clinical knowledge, generated through human perception. In clinical practice, new data are usually interpreted through similarity-based reasoning, combining both visual features and semantic concepts. (André et al., 2012a; André et al., 2012c) defined 8 mid-level binary semantic concepts that were either present or not in a colonic endomicroscopic video sequence. A Fisher-based approach was utilised to estimate the expressive power of each of the visual words (estimated as per (André et al., 2011b)) to each of these 8 semantic concepts. The derived semantic signatures were found to be informative and consistent with the low-level visual features, providing some relevant semantic translation, more familiar to the clinicians' own language, of the visual retrieval outputs. In a separate attempt to alleviate the semantic gap, (Watcharapichat, 2012) proposed an interactive approach that the user has the ability to provide "relevance feedback" on the previously retrieved content, enabling the system to iteratively improve upon the search results. The feedback was combined on Isomap dimensionality reduction for improved performance and efficiency. (Tous et al., 2012) developed a multimedia retrieval software enabling querying via low-level, image-based features as well as high level key-word semantic descriptions. The software ensured compatibility with third party applications through interface compliance with the MPEG Query Format (ISO/IEC 15938-12:2008) and JPEG Search (ISO/IEC 24800) standards.

In an attempt to improve the retrieval performance of (André et al., 2011b), (Kohandani Tafresh et al., 2014) introduced a simple and efficient semi-automated approach allowing clinicians to create more meaningful queries than unprocessed endomicroscopic video sequences. The approach automatically temporally segmented endomicroscopic video sequence based on kinematic stability assessment, with informative sub-segments assumed spatially stable. Then, the clinician could manually select stable sub-sequences of interest generating a new augmented query video, leading to more reproducible and consistent retrieval results. (Gu et al., 2017) proposed Unsupervised Multimodal Graph Mining (UMGM), a framework mining the latent similarity amongst endomicroscopic mosaics and histology patches for enhanced CBIR performance. While an extension of (Gu et al., 2016), UMGM employed graph-based analysis over a large collection of histology patches without supervised information (matching pairs), minimising latent space distance between similar pairs while maximising the distance between dissimilar pairs.

## 6. Limitations and opportunities

Fibre bundle based endomicroscopy (FBEμ) offers several enabling capabilities for diagnostic and interventional procedures in a range of clinical indications. The literature to date has established a solid understanding of the limitations inherent to imaging through coherent fibre bundles, making substantial progress in terms of associated image computing methodologies. Characteristic examples of concentrated research effort have been (i) compensating for the honeycomb effect through the irregular, sparse sampling introduced along the coherent fibre bundle, and (ii) extending the limited field of view, a direct consequence of the fibre bundle miniaturisation for guidance through an endoscope's working channel, through mosaicing spatially adjacent frames. Yet, FBEμ is a still a fledging imaging technology with tremendous potential for improvement assuming the research/technical challenges can be overcome. Throughout this review the following major image computing challenges/opportunities have been identified.

**Table 6.** Overview of image retrieval approaches for fibred endoscopic imaging.

| Topic | References | Methodology | Comments |
|---|---|---|---|
| Image retrieval through low-level visual features. | (André et al., 2009a) | Bag of Visual Words (k-means clustering) of multi-scale SIFT descriptors extracted from regularly distributed circular regions. | Thorough methodologies for image and video retrieval based solely on low-level information extracted from images. Due to lack of relevant ground truth, methodologies were evaluated as binary classification tasks (instead of retrieval). |
| | (André et al., 2009b) | Introduce (i) spatial information between local features by exploiting the co-occurrence matrix of their visual words (ii) temporal relationship across frames through mosaicing. | |
| | (André et al., 2010) | Deriving visual words from individual frames and weighting the contributions of local regions through the relevant overlap rate derived during mosaicing. | |
| | (André et al., 2012b; André et al., 2011b) | Combining and clinically testing above approaches as a binary classification (kNN) between neoplastic/benign colonic epithelium. | |
| | (André et al., 2011a) | (i) Generate the "perceived similarity" ground truth (manual assessment – Likert scale), and (ii) learn an adjusted similarity/distance metric (linear transform) for optimal mapping of video signatures (histograms of visual words). | First attempt to evaluate directly the performance of endomicroscopic video retrieval, through generating the perceived similarity of ground truth. |
| Image retrieval combining low-level visual features with high-level semantic context. | (André et al., 2012a; André et al., 2012c) | Fisher-based approach transforming visual word histograms to 8 binary semantic concepts. Combine with adjusted similarity distance to improve "perceived similarity". | Bridging the semantic gap between low-level visual features, extracted from the images, and high-level clinical knowledge, generated through human perception. |
| | (Watcharapichat, 2012) | Gabor filter and Earth Mover's Distance based retrieval enhanced through iterative "relevance feedback" and Isomap dimensionality reduction. | |
| | (Tous et al., 2012) | Retrieval via (i) low-level, image-based features (LBPs & k-NN with Euclidian or Manhattan distances), (ii) high level key-word semantic descriptions (Apache Lucene search engine), and (iii) third party software compatibility through MPEG Query Format & JPEG Search standards. | |
| Other image retrieval approaches | (Kohandani Tafresh et al., 2014) | Semi-automated query adaptation of (André et al., 2011b) via (i) temporal segmentation based on kinetic stability (Euclidean distance of SHIFT descriptors across consecutive frames), and (ii) manual selection of spatially stable segments. | Adaptations of (André et al., 2011b) enhancing retrieval performance. |
| | (Gu et al., 2017) | Unsupervised, multimodal graph mining (i) deriving similar (cycle consistency) and dissimilar (geodesic distance) FBEµ and histology frame pairs, (ii) learning discriminative features in the associated latent space. | |

## 6.1. Image reconstruction

Image reconstruction research has concentrated predominantly in compensating for the honeycomb effect on raw FBEµ images, a consequence of the sparse, irregular sampling through the coherent fibre bundle. Yet, even straightforward approaches such as bilinear interpolation between the cores, as currently used in clinical practice (Cellvizio, Mauna Kea Technologies), have been found to generate satisfactory results, with subsequent improvements perceived as predominantly aesthetic. In contrast, very limited research has been performed compensating for other inherent artefacts known to have limiting effect on the imaging capabilities of the technology, such as (i) variable coupling and background response (due to irregularities amongst cores physical properties) and (ii) inter-core coupling across neighbouring cores. Optimal solutions to these problems can have a direct impact on the imaging signal to noise ratio, contrast and potentially the spatial resolution (computationally supressing cross coupling can conceivably enable smaller inter-core distances). Furthermore, with the notable exception of the work by (Vercauteren et al., 2013), there have been no studies on multi-colour data acquisition, investigating and compensating for the effect of the aforementioned coupling/background artefacts along with other inherent limitations such as spectral mixing. Such enhanced imaging capabilities are of paramount importance into the advancing molecular endomicroscopy which has stringent requirements in terms of light detection (preferably at multiple wavelengths), especially when imaging small targets such as bacteria superimposed upon highly fluorescent background structures. Finally, existing reconstruction approaches tend to concentrate on a single limitation in FBEμ imaging, intrinsic to the coherent fibre bundle characteristics, either ignoring or downplaying the relevance of other limitations in the reconstructed images. In real-world applications this is rarely the case. There is therefore scope for the development of a unified image reconstruction methodology that compensates for a range of limitations, including but not limited to irregular sampling, varying coupling efficiency and inter-core coupling along with additional challenges introduced in multi-spectral acquisition such as chromatic aberrations and spectral mixing. This is of greater importance to widefield FBEμ, where poor sectioning already reduces limits of detection and subsequently the imaging capabilities of the technology. The emergence of deep-imaging (Wang, 2016), employing data-driven deep learning (Convolutional Neural Networks) for image formation/reconstruction from raw, irregularly sampled data, is expected to generate tremendous opportunities in biomedical imaging in general and FBEμ in extension. (Ravì et al., 2018) for example has recently demonstrated a deep-learning based super-resolution pipeline for FBEμ. Yet, this direction, while very promising, will eventually lead to additional challenges regarding the need for large amounts of carefully chosen, and meaningful "gold-standard" data to form the basis of the learning and inference processes. Furthermore, convolution filers have been the cornerstone of state-of-the-art deep learning approaches for classical regularly sampled images. Yet, FBEμ images are sparsely and irregularly sampled through a coherent fibre bundle subsequently reconstructed to a regularly sampled image, potentially introducing uncertainty to the image reconstruction process. There is therefore scope for developing novel deep-learning architectures applied directly on the irregularly sampled data.

### 6.2. Pathology detection and quantification

There is a substantial body of work on the classification of frames into clinically relevant groupings, based predominantly on the binary classification between healthy and pathological frames over a range of organ systems and associated pathologies. Generating hand-crafted feature descriptors and training a binary or multi-class classifier has been shown to generate reliable results in parsing videos and detecting abnormalities in endomicroscopic frame sequences. Yet, there has been very limited work on the semantic segmentation and subsequent pathology detection and quantification for FBEμ frames and mosaics. (IIzadyyazdanabadi et al., 2018) for example proposed a weakly supervised CNN architecture for localising brain tumours in eCLE images. Pathology quantification will be imperative to any viable Computer-aided detection (CADe) and Computer-aided diagnosis (CADx) system. Furthermore, the existing image quantification studies have primarily adopted, empirical, ad hoc methodologies along with heuristic parameter estimation with hard thresholds, tested on very limited data. As a result, this can lead to poor generalisation as well as limited clinical utility. There is therefore an opportunity and need for the development of customised and robust methods that analyse and quantify the contents of FBEµ images, which when combined with state of the art detection and classification approaches, can identify and quantify pathology as the cornerstone for invaluable CAD systems. Ultimately, in certain clinical applications, pathology detection and quantification will also be aided by targeted molecular imaging agents.

## 6.3. Integration

To date, the tasks of image reconstruction, analysis and understanding have been dealt with independently, with notable exceptions the works of (Hu et al., 2010; Ravì et al., 2018; Vercauteren et al., 2006) that employ image mosaicing techniques to generate a super-resolved reconstructed image. Consequently, image reconstruction has been optimised primarily for user experience. While user experience in a clinical setting is an extremely important factor, contributing to the success of the endomicroscopic procedure through effective guidance and on-target sampling, it is not necessarily a primary concern during the automated detection and quantification of pathology. Moreover, most studies have employed unimodal information derived exclusively from endomicroscopic images, with a small number of multimodal attempts integrating histological (Gu et al., 2017; Gu et al., 2016), demographic and clinical (Seth et al., 2016) information in the decision-making pipeline. Yet, endomicroscopy (predominantly due to limited FOV and guidance capabilities) is unlikely to be used as a stand-alone tool in the clinical workflow. FBEµ will be integrated as part of a multimodality approach consolidating imaging across a range of scales, from organ level (radiology) to cellular level (microscopy), along with other clinically relevant information. There is therefore scope for (i) incorporating multi-modal information in the decision-making algorithms, and (ii) integrating the reconstruction, analysis and understanding of endomicroscopic images to novel unified frameworks with joint loss functions, optimised for the task in question such as identifying and quantifying pathology.

## 6.4. Data availability

Recent developments in Convolutional Neural Networks (CNNs) have acted as vehicle to substantial advances to image analysis and understanding across an ever-increasing range of areas, including medical imaging, with applications in image reconstruction, classification, segmentation and registration. Yet, to date there have been just a limited number studies employing Convolutional Neutral Networks for the classification and retrieval on FBEµ frames and mosaics. Instead, image understanding tasks has been tackled predominantly through traditional machine learning pathways, defining hand-crafted feature descriptors and subsequently training a binary or multi-class classifier on this feature set. A key constraint in the effective adaptation and adoption of the technology (CNNs) has been, to a large extent, the limited data and associated annotations available. In particular, the plurality of FBEµ classification/retrieval studies have employed limited data, ranging between 100 and 200 annotated frames for combined training, validation and testing, with several studies using datasets of less than 100 frames. Furthermore, the available data have for the most part been acquired from a single clinical site and many times from a single operator, introducing potential bias and hindering the ability of the proposed methodologies for widespread generalisation. Similarly, there is often a lack of a gold reference standard and manual annotations can be weak, demonstrating large inter- and intra-operator variability. In tasks such as image restoration and analysis, the proposed methodologies of assessment have been constrained to simple simulated data, test targets, and in some cases to a very limited number of biological samples. There is therefore a need for the development of (i) large data depositories, containing a diverse collection of frames sequences acquired from different operators at multiple sites across the world, with easy access for the endomicroscopy research community, (ii) associated manual annotations, ideally from multiple operators with varied level of expertise, with quantifiable inter- and intra-operator variability. Providing standardised annotation tools, available alongside the data depositories can further enhance the consistency and robustness of these annotations.

### 6.5. Real-time capability

In much of the FBEμ image computing literature to date, the proposed methodologies have limited or no capacity for real-time application. Given the potential for FBEμ to perform *in vivo*, *in situ* assessment, at microscopic level (optical biopsies), the lack of real-time capability impairs the clinical application for such algorithms. There is therefore a necessity to design and test methodologies, from the ground up, with particular consideration for their real-time potential under pragmatic computational resources (at the time of testing and near future) for the intended clinical application.

## 7. Conclusions

Fibre bundle based endomicroscopy (FBEμ) is a relatively new medical imaging modality. Yet, the real time, microscopic imaging capabilities, commonly referred to as optical biopsy, make FBEμ a very promising diagnostic and monitoring tool, particularly when combined in the future with molecular imaging agents. Imaging through a miniaturised coherent fibre bundle, typically guided to the region of interest through the working channel of an endoscope, imposes a number of inherent limitations to the technology. These limitations have motivated a diverse and ever-growing area of research for tailored image computing solutions. To date, considerable progress has been made in (i) image reconstruction, compensating for the honeycomb introduced by the coherent fibre bundle, (ii) extending the limited field of view through mosaicing adjacent frames, and (iii) classifying frames amongst two or more clinically relevant categories. However, there are still significant research challenges and opportunities remain for FBEμ to realise its full clinical potential.

## Acknowledgements

Funding: This work was supported by the Engineering and Physical Sciences Research Council (EPSRC, United Kingdom) [EP/K03197X/1 and NS/A000050/1], as well as the Wellcome Trust [203145Z/16/Z and 203148/Z/16/Z].

## Declaration of interest

Professor Vercauteren is a shareholder of Mauna Kea Technologies (Paris, France). Professor Dhaliwal is founder and shareholder of Edinburgh Molecular Imaging (Edinburgh, UK) and has in the past received funds for travel and meeting attendance from Mauna Kea Technologies (Paris, France).